\documentclass[a4paper,10pt]{article}
\pdfoutput=1
\usepackage{amssymb}
\usepackage{geometry}
\usepackage[usenames,dvipsnames]{color}
\usepackage{slashed}
\usepackage[table]{xcolor}
\usepackage{graphicx}
\usepackage{mathtools}
\usepackage{cite} 
\usepackage{hyperref}

\usepackage{amsthm}
\usepackage{bm}

\theoremstyle{plain}

\theoremstyle{definition}

\theoremstyle{remark}

\title{\bf{Partial local entropy and anisotropy in deep weight spaces}}
\date{}
\author{Daniele Musso\footnote{daniele.musso@usc.es, daniele.musso@cesga.es, mudaniele@yahoo.com}}

\begin{document}

\maketitle
\vspace{-20pt}
\begin{center}\it{
Departamento de  F\'\i sica de Part\'\i  culas,\\
Universidade de Santiago de Compostela (USC)\\
\vspace{5pt}
Instituto Galego de F\'\i sica de Altas Enerx\'\i as (IGFAE)\\
E-15782 Santiago de Compostela, Spain\\
\vspace{5pt}
and\\
\vspace{5pt}
Inovalabs Digital S.L. (TECHEYE),\\
E-36202 Vigo, Spain
\vspace{5pt}
and\\
\vspace{5pt}
Centro de Supercomputaci\'on de Galicia (CESGA),\\
s/n, Avenida de Vigo, 15705 , Santiago de Compostela, Spain}
\end{center}
\vspace{5pt}

\begin{abstract}
We refine a recently-proposed class of local entropic loss functions by restricting the smoothening regularization to only a subset of weights. 
The new loss functions are referred to as \emph{partial} local entropies.
They can adapt to the weight-space anisotropy, thus outperforming their 
isotropic counterparts. We support the theoretical analysis with experiments on image classification tasks
performed with multi-layer, fully-connected and convolutional neural networks. 
The present study suggests how to better exploit the anisotropic nature of 
deep landscapes and provides direct probes of the shape of the minima encountered by 
stochastic gradient descent algorithms.
As a by-product, we observe an asymptotic dynamical regime at late training times where the temperature of all the 
layers obeys a common cooling behavior.

\end{abstract}
\newpage
\tableofcontents

\section{Introduction}

Recent studies on the weight space of deep neural networks \cite{Baldassi2015SubdominantDC,2016arXiv161101838C}
have highlighted the existence of rare subdominant clusters of configurations which yield a high test accuracy.
Although these clusters constitute a deviation from typicality, they are efficiently encountered by 
\emph{stochastic gradient descent} (SGD) algorithms and correspond to wide valleys of suitable 
loss functions, such as cross entropy \cite{Baldassi_2019}.

An analogous circumstance occurs in the context of constraint satisfaction problems, 
where the chase after clusters of solutions is improved when the 
loss function gets supplemented by a term that encourages a local high density of solutions \cite{Baldassi2016LocalEA}. 
In order to find the number of solutions contained in a vicinity of a specific weight configuration, one can 
define a local solution-counting functional, namely, a \emph{local entropy}.

Classification tasks performed by means of quantized neural networks (where the weights are discrete)
can be interpreted as constraint satisfaction problems. There are however two reasons to generalize 
the concept of local entropy: First, classification problems are typically required to reach a high but not 
necessarily perfect accuracy; second, they are often approached with machines that have continuous weights.%
\footnote{Up to the numerical precision employed.}
The strict counting of solutions employed for constraint satisfaction problems can therefore be relaxed to just an incentive which encourages 
a high local density of high-accuracy configurations. A local averaging of 
the loss, for instance,  is expected to have such an effect, but other deformations of the loss 
yielding a local smoothening can be valid choices too.

A specific smoothening procedure of the loss function can be enforced 
by means of a spatial convolution with an Euclidean heat kernel, whose spread is controlled by a parameter $\gamma$,
\begin{equation}\label{loc_ent}
 {\cal F}(\beta, \gamma; \boldsymbol W) = -\log \int d^N W'\ \exp\left(-\beta {\cal L}(\boldsymbol W') - \frac{\gamma}{2}||\boldsymbol W-\boldsymbol W'||^2_2\right) \ ,
\end{equation}
where both $\boldsymbol W$ and $\boldsymbol W'$ parametrize the $N$-dimensional weight space,
$||.||_2$ represents the Euclidean norm and ${\cal L}$ is a generic loss function;
adopting an energetic interpretation of the loss, 
the parameter $\beta$ corresponds to an inverse temperature.
In the limit $\beta \rightarrow 0$, the integral in \eqref{loc_ent} can be interpreted as
(the continuum version of) a weighted counting of the configurations $\boldsymbol W'$ where the weighting decreases 
exponentially with their distance from $\boldsymbol W$ \cite{Baldassi_2016}.

The smoothening introduced by \eqref{loc_ent} is \emph{isotropic} in weight space. 
However, when optimizing with SGD, the gradient noise depends in general on both the position and the direction,
this being actually a key factor for the success of SGD algorithms \cite{Xie2020ADT}. 
Therefore, it is natural to expect that a refinement of the smoothening functional able to suitably exploit 
the anisotropy of gradient noise can significantly improve its regularizing effects. Besides, such refinement
can furnish an interesting new probe of the weight space.

The present paper focuses on \emph{partial}, entropic and local smoothening, namely a smoothening analogous to \eqref{loc_ent}
applied to just a subsets of weights.
This allows one to address weight-space anisotropy in a direct and active way. We will loosely 
adopt the term \emph{partial local entropy} to convey this idea irrespective of the details of 
the particular smoothening technique, 
as long as it corresponds to an incentive to local high density of high-accuracy configurations
restricted to a subset of weights.%
\footnote{The functional ${\cal F}(\beta, \gamma; \boldsymbol W)$ defined in \eqref{loc_ent} 
can be interpreted in analogy to a thermodynamical potential; as such, it should be referred 
to as local \emph{free} entropy, this extra connotation is sometime omitted to avoid clutter.}

\section{Anisotropy in weight space}

By definition the neurons of a deep network are arranged on different layers
and such architecture imposes a natural hierarchy among them, according 
to their depth within the network. In a fully-connected setting, the receptive 
field of each neuron coincides with the whole input, 
however deeper neurons are fed with signals that have been 
pre-processed by lower-lying neurons. Roughly, while the neurons in the 
first layer compute a weighted sum of the network inputs, the neurons
in the second layer compute a weighted sum of the outputs of the first 
layer, that is, a weighted sum of a weighted sum of the network inputs.
Such compositional nature of the operation performed by each subsequent layer
suggests that the depth of the network corresponds to a hierarchy in 
combinatorial complexity \cite{Wei}.%
\footnote{One can rephrase such combinatorial complexity 
in terms of correlations among the input channels: the neurons in the first 
layer are sensitive to the inputs individually, so they respond to 1-point 
correlations; the neurons belonging to the $n$-th layer, instead, 
are sensitive to $n$-point correlations, that is, the joint correlations
of $n$ inputs.}
Any isotropic assumption about the 
weight space neglects this structural hierarchy, thereby it should be  
regarded with caution if not even suspicion.

Careful consideration of the hierarchical anisotropy of the weight space has led to important 
insight about the inner workings of neural networks (also in the biological domain \cite{Poggio}) as well as 
improvements in the optimization of artificial neural networks.%
\footnote{To this regard, two relevant examples are Kaiming
weight initialization \cite{he2015delving} and regularization 
by means of anisotropic noise injection \cite{2018arXiv180300195Z,Musso:2020itr}.}
Gradient noise depends on both position and direction and 
its covariance matrix is correlated to the Hessian matrix of the loss function, which makes SGD  
escape exponentially fast from sharp minima \cite{Xie2020ADT}.%
\footnote{In order to maintain the analysis as simple as possible, 
in the present paper we do not exploit the Hessian matrix of the loss function
to define specific partial local entropies, yet this represents 
an interesting direction for further investigation. Specifically, information about 
the eigenvalues of the Hessian matrix could be useful in \emph{scoping} the hyper-parameter 
$\gamma$ (see \eqref{loc_ent}), namely, in adjusting its value during optimization in an adaptive fashion.}
Thus, it is 
fair to consider weight-space anisotropy as one of the main features 
at the root of the effectiveness of SGD algorithms in reaching high test 
accuracy and generalization.

\subsection{Layer temperature and asymptotic cooling}
\label{asy_coo}

The learning dynamics of a deep neural network trained with SGD is in general a complex process.
The system is out of equilibrium and, given the dependence of the gradient noise on the position in weight space,
one cannot schematize the training as the evolution of a system in contact with an equilibrium 
thermal reservoir. Nonetheless, it is still possible to define a temperature as the variance 
of the gradient noise when schematizing the training evolution in terms of a Brownian motion \cite{Wei,DBLP:journals/corr/abs-1710-11029,Musso:2020itr}. 
More precisely, one has to focus on the covariance matrix $D(\boldsymbol W)$ characterizing the 
stochastic Wiener process.%
\footnote{We refer to \cite{DBLP:journals/corr/abs-1710-11029} for the definition of the covariance 
matrix $D(\boldsymbol W)$. The analysis of a Brownian motion by means of the Fokker-Planck equation 
encodes both the noise anisotropy and its dependence on position through the covariance matrix $D(\boldsymbol W)$
\cite{DBLP:journals/corr/abs-1710-11029,DASILVA200416}.}

Let us focus for a moment on a specific point $\boldsymbol W^*$ in weight space. Given the anisotropy of $D(\boldsymbol W^*)$, 
it is impossible to define a unique temperature characterizing all directions, but one can in principle 
still define a temperature for each direction. Since we are working in a space with very high dimensionality,
this is hardly of any help. However, we should recall that there is a natural grouping of the directions in weight space 
provided by the layered structure of the network. Furthermore, it is possible to define layer variables which average over
the weights belonging to the same layer. One can consider fluctuations of such layer variables 
that, due to the averaging over a layer, are expected to be stabler and reflect the hierarchy 
of the architecture. Accordingly, one can define a layer temperature corresponding to the variance 
of such layer averaging of gradients.%
\footnote{We underline that a direct analysis of the variance of the gradient noise for the single weights shows that 
in general the weights belonging to the same layer can \emph{not} be characterized by a common 
temperature. 
Said otherwise, the possibility of defining a layer temperature does 
not imply thermal isotropy within the subspace spanned by the weights of the same layer.} This corresponds to regarding the layers as if they were the individual 
units of a neural network; despite being a crude approximation, this 
could help gaining useful insight about the training dynamics \cite{DBLP:journals/corr/Shwartz-ZivT17}.%
\footnote{Reference \cite{DBLP:journals/corr/Shwartz-ZivT17} has been debated in the subsequent literature,
we thank the referee for stressing this point and for suggesting a wider set of references
useful for a critical analysis \cite{michael2018on,goldfeld2019estimating}.}

\begin{figure}
 \begin{center}
  \includegraphics[width=0.99\textwidth]{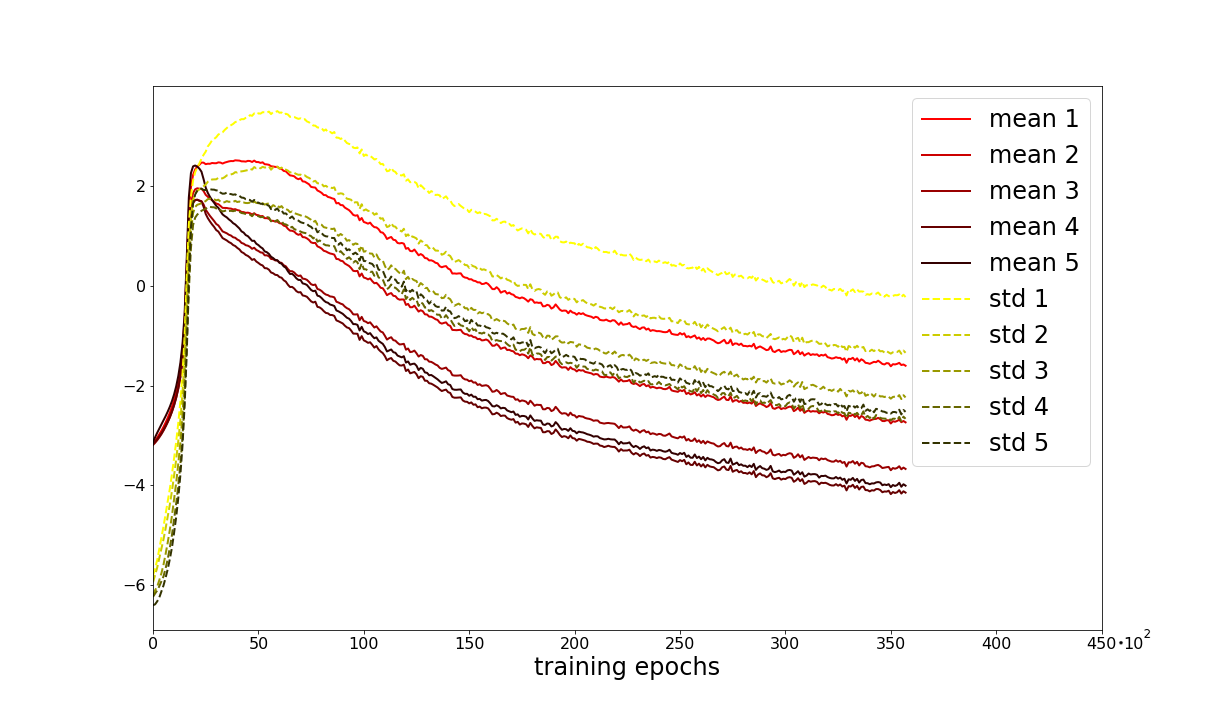}
  \includegraphics[width=0.99\textwidth]{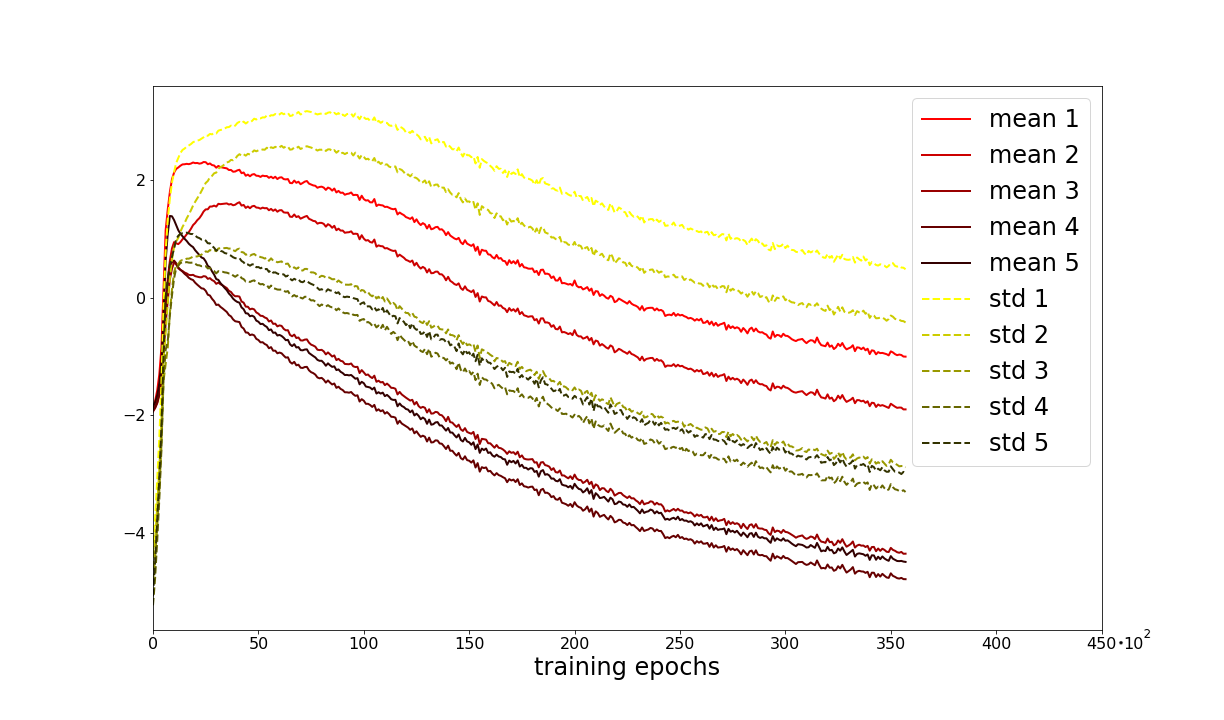}
 \end{center}
 \caption{The training signal $s_I$ defined in \eqref{tra_sig}
 --where $I\in \{1,2,3,4\}$ labels the layers of the network-- is represented with solid lines;
 the dashed lines represent the associated standard deviations. The plots depict a long training of two
 4-layer fully-connected neural networks on MNIST with either ReLU (top plot) or TanH (bottom plot) activation functions. 
 Three distinct dynamical regimes emerge in both plots: 1) an early signal-dominated
 regime; 2) a short, intermediate and noise-dominated regime; 3) an eventual, long and noise-dominated regime 
 where all quantities decay sub-exponentially with a common behavior (the vertical axes are natural logarithms).}
 \label{ddt}
\end{figure}

The layer temperature is a characterization of the noise of the training signal $s_I$ through layer $I$, defined as
\begin{equation}\label{tra_sig}
 s_I = \frac{1}{N_I} \sum_{\omega\in \Omega_I} ||\nabla_{\omega} {\cal L}(\boldsymbol W)||_2\ ,
\end{equation}
where $\Omega_I$ denotes the set of the $N_I$ weights connecting the $I$-th layer with its inputs,
$||.||_2$ represents the Euclidean norm and ${\cal L}(\boldsymbol W)$
is the loss evaluated at the weight configuration $\boldsymbol W$. 
The training signals $s_I$ and their noise evolve during optimization and it is possible 
to isolate different regimes in the training dynamics. In \cite{DBLP:journals/corr/Shwartz-ZivT17}
the authors observed that a possibly generic dynamic transition occurs when the signal$/$noise ratio switches
from being initially dominated by the signal to being later dominated by noise. This occurs quite abruptly (in terms of 
optimization time) and approximately at the moment when the training signal attains its maximum value, see Figure \ref{ddt}.

The numerical studies that we performed suggest the generic presence of a further dynamic transition,
occurring at later stages of the training. This eventual regime is characterized by a sub-exponential decay 
of both signal and noise for all layers. Interestingly, the sub-exponential contraction of the signal and the noise 
for all the layers is characterized by a common decaying behavior. At late times, the hierarchy between layers is therefore 
preserved and gets frozen: the dynamics of all the layers can in fact be described factorizing the common sub-exponential decay.%

Interpreting the noise as a temperature and adopting a renormalization group language, 
the eventual sub-exponential cooling (possibly turning exponential at asymptotically late times)
is suggestive of an infrared fix point, where quantities evolve by a common rescaling without distortion at asymptotic low energies.%
\footnote{Here it emerges a potential connection to studies of neural networks under the perspective of scaling rules, 
see for instance \cite{2017arXiv171006451S,sharma2020neural}.} It is relevant to stress 
that Figure \ref{ddt} has been obtained \emph{without} adopting weight-decay regularization.
Moreover, we have obtained qualitatively similar results both with ReLU and TanH activation functions;
while the former is scale covariant, the latter is not.

As already stressed, even if the layer-wise account gives a very coarse-grained picture of the actual training dynamics, 
still it confirms the importance of anisotropy throughout the whole training process, including at asymptotic late times 
where the in-sample loss and the test error have long stabilized.

\section{Partial local free entropy}
\label{PLE}

For the sake of generality, the present section is rather technical. The reader who is just 
interested in the specific losses used in the experiments can jump to Section \ref{exp_MF} and focus on 
the loss functions \eqref{PLEA} and \eqref{PLA} without missing the core ideas.
\vspace{4pt}

We consider the cross-entropy loss ${\cal L}_{\text{c.e.}}(\boldsymbol W)$ as the baseline function to be smoothened;
$\boldsymbol W$ is a vector indicating a configuration in weight space. We consider $y$ additional configurations 
$\boldsymbol W + \Delta \boldsymbol W^a$ with $a=1,...,y$, shifted by a uniformly distributed random vector $\Delta \boldsymbol W^a$. 
The loss corresponding to each configuration is supplemented by an additional term measuring its distance from 
the unperturbed point $\boldsymbol W$. For the moment we let the distance function $d_{R,k}(\Delta \boldsymbol W^a)$
be arbitrary but we assume it depends on two parameters, to be specified later.%
We consider the new loss
\begin{equation}\label{smu}
 {\cal M}(R, k, y; \boldsymbol W) \equiv -\log \left\{
 \frac{1}{y+1}\left[e^{-{\cal L}_{\text{c.e.}}(\boldsymbol  W)}
 + \sum_{a=1}^y e^{-{\cal L}_{\text{c.e.}}(\boldsymbol  W+\Delta \boldsymbol W^a) - d_{R,k}(\Delta \boldsymbol W^a)}\right]\right\}\ ,
\end{equation}
normalized with respect to the number of sampling points $y+1$. Roughly, the loss ${\cal M}$ amounts to the logarithm 
of an average of exponentials. In the case of just one sampling point, $y=0$, ${\cal M}$ coincides with the baseline loss,
\begin{equation}
 {\cal M}(R, k, y=0; \boldsymbol W) = {\cal L}_{\text{c.e.}}(\boldsymbol  W)\ .
\end{equation}

\begin{figure}
 \begin{center}
  \includegraphics[width=0.49\textwidth]{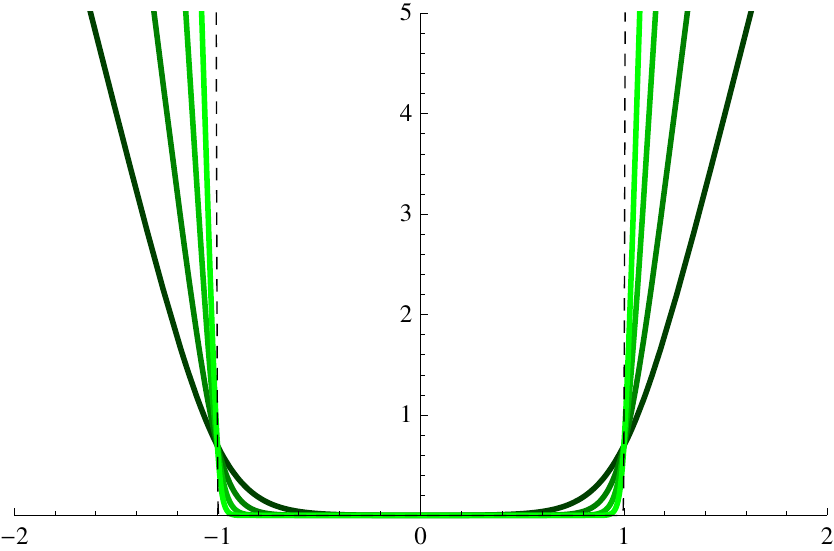}
  \includegraphics[width=0.49\textwidth]{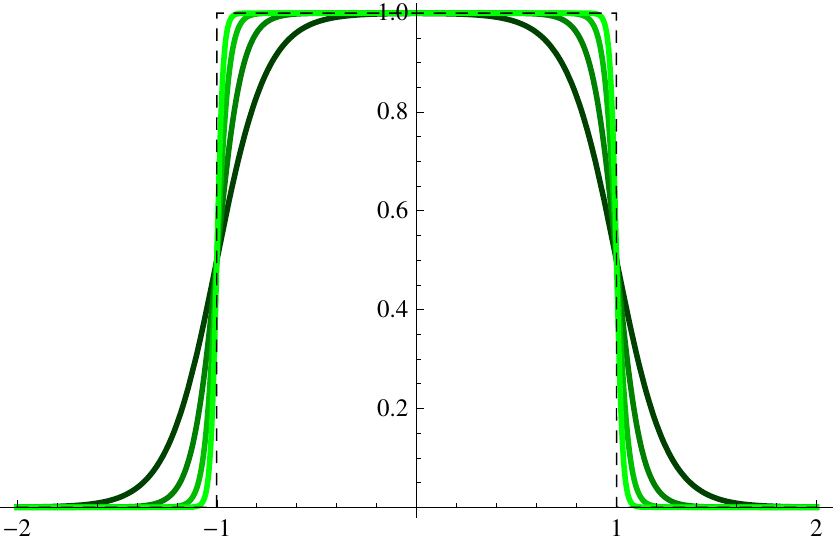}
 \end{center}
 \caption{1-dimensional section of the distance function $d_{R,k}$ defined in \eqref{dista} (left plot)
 and of the kernel $K_{R,k}$ defined in \eqref{kernel} (right plot); in the plots $R=1$ and $k=2^2,2^3,2^4,2^5$
 from darker to lighter.} 
 \label{sca_buc}
\end{figure}

We choose the following distance function
\begin{equation}\label{dista}
 d_{R,k}(\Delta \boldsymbol W) \equiv 
 - \log \prod_{i=1}^N \left[\left(1-\frac{1}{1+e^{-2 k (\Delta W_i - R)}}\right)\frac{1}{1+e^{-2 k (\Delta W_i + R)}}\right]\ ,
\end{equation}
which depends on two real parameters, $R$ and $k$. In the $k\rightarrow \infty$ limit, the kernel 
\begin{equation}\label{kernel}
 K_{R,k}(\Delta \boldsymbol W) \equiv e^{-d_{R,k}(\Delta \boldsymbol W)}\ ,
\end{equation}
reduces to the characteristic function of the $N$-dimensional hyper-cube $H_{\boldsymbol W, R}$ centered in $\boldsymbol W$ 
with edges $2R$ long,%
\footnote{Recall that the Heaviside step function $\Theta(x)$
can be obtained as the limit of infinite sharpness for a sigmoid function, namely
\begin{equation}
 \Theta(x) = \lim_{k \rightarrow +\infty} \frac{1}{1+e^{-2kx}}\ .
\end{equation}}
\begin{equation}
 \lim_{k\rightarrow +\infty } K_{R,k}(\Delta \boldsymbol W) = 
 \prod_{i=1}^N \left[1-\Theta(\Delta W_i - R)\right] \Theta(\Delta W_i + R)\ .
\end{equation}
Thus, the parameter $R$ represents the effective linear size of the support of the kernel \eqref{kernel},
while $k$ controls its sharpness, see Figure \ref{sca_buc}. In the infinite sharpness limit, $k\rightarrow \infty$, 
the random displacement vectors $\Delta \boldsymbol W^a$ in \eqref{smu} are sampling the hyper-cube $H_{\boldsymbol W, R}$ uniformly.

Taking an infinite number of sampling points,
\begin{equation}\label{inf_cop}
 {\cal M}(R, k, y; \boldsymbol W) \xrightarrow[y\rightarrow +\infty]{}\ {\cal F} (R, k; \boldsymbol W)\ ,
\end{equation}
where
\begin{equation}\label{fam}
 {\cal F} (R, k; \boldsymbol W) \equiv - \log \int d^N W' \ e^{-{\cal L}_{\text{c.e.}}(\boldsymbol W')}\ K_{R,k}(\boldsymbol W'-\boldsymbol W) \ ,
\end{equation}
defines a parametric family ${\cal F} (R, k; \boldsymbol W)$ of \emph{local free entropies}, in analogy with \eqref{loc_ent}.%
\footnote{The particular local free entropy specified in \eqref{loc_ent} is associated to a different choice of distance,
namely 
\begin{equation}
 d(\gamma; \Delta \boldsymbol W) = \gamma || \Delta \boldsymbol W||_2^2\ .
\end{equation}
}
Taking the $k\rightarrow \infty$ limit of \eqref{fam}, one obtains
\begin{equation}
  \lim_{k \rightarrow + \infty} {\cal F} (R, k; \boldsymbol W) = 
  - \log \int_{H_{\boldsymbol W, R}} d^N W'\ e^{-{\cal L}_{\text{c.e.}}(\boldsymbol  W')} \ .
\end{equation}

To recapitulate, in the limit of large number of sampling points, $y\rightarrow \infty$,
the loss function ${\cal M}(R, k, y; \boldsymbol W)$ approximates a parametric family of 
\emph{free local entropy} functions \eqref{fam} parametrized by the effective linear size $R$ of the
smoothening region (in weight-space) and the sharpness $k$ of the associated kernel \eqref{kernel}.

In order to define \emph{partial local free entropies} we have just to generalize the passages above to the case 
where only a subset of weights is smoothened over. We can define a discrete indicator function $\boldsymbol U$ taking 
values in $\{0,1\}^N$ and defined on the $N$ dimensions of weight space: it takes value $1$ on the directions along 
which we smoothen the loss, and $0$ on the remaining directions in weight space. Thinking to 
$\boldsymbol U$ as an $N$-dimensional vector, it provides an un-normalized projector onto the subset of weights
considered for smoothening.
We can thus define a restricted version of the distance function $d_{R,k}(\Delta \boldsymbol W)$,
\begin{equation}\label{res_dis}
 d_{R,k}^{[\boldsymbol U]}(\Delta \boldsymbol W) \equiv d_{R,k}\Big((\Delta \boldsymbol W \cdot \boldsymbol U)\, \boldsymbol U\Big)\ ,
\end{equation}
where $\cdot$ indicates the scalar product of $\mathbb R^N$ in the $N$-dimensional weight space.

Adopting the restricted distance \eqref{res_dis}, we can repeat the same steps as above:
first consider
\begin{equation}\label{smu_U}
 {\cal M}^{[\boldsymbol U]}(R, k, y; \boldsymbol W) \equiv -\log \left\{
 \frac{1}{y+1}\left[e^{-{\cal L}_{\text{c.e.}}(\boldsymbol  W)}
 + \sum_{a=1}^y e^{-{\cal L}_{\text{c.e.}}(\boldsymbol  W+\Delta \boldsymbol W^a) - d_{R,k}^{[\boldsymbol U]}(\Delta \boldsymbol W^a)}\right]\right\}\ ,
\end{equation}
then take the $y\rightarrow \infty$ limit 
\begin{equation}\label{inf_cop_U}
 {\cal M}^{[\boldsymbol U]}(R, k, y; \boldsymbol W) \xrightarrow[y\rightarrow +\infty]{}\ {\cal F}^{[\boldsymbol U]} (R, k; \boldsymbol W)\ ,
\end{equation}
where
\begin{equation}\label{fam_U}
 {\cal F}^{[\boldsymbol U]} (R, k; \boldsymbol W) \equiv - \log \int d^N W' \ e^{-{\cal L}_{\text{c.e.}}(\boldsymbol W')}\ 
 K_{R,k}^{[\boldsymbol U]}(\boldsymbol W'-\boldsymbol W) \ ,
\end{equation}
represents a parametric family of \emph{partial local free entropies}.
Eventually, take the $k\rightarrow \infty$ limit, 
\begin{equation}
 {\cal F}^{[\boldsymbol U]}(R; \boldsymbol W) \equiv \lim_{k \rightarrow + \infty} {\cal F}^{[\boldsymbol U]}(R, k; \boldsymbol W)\ ,
\end{equation}
where
\begin{equation}
 {\cal F}^{[\boldsymbol U]}(R; \boldsymbol W) \equiv - \log \int_{H_{\boldsymbol W, R}^{[\boldsymbol U]}} 
 d^N W'\ e^{-{\cal L}_{\text{c.e.}}(\boldsymbol  W')} \ ;
\end{equation}
the integration region $H_{\boldsymbol W, R}^{[\boldsymbol U]}$ is a hyper-cube extended only in the 
directions along which $\boldsymbol U$ is non-null.

\subsection{A simpler entropic loss}

It is interesting to seek for a simpler loss which could somehow preserve the smoothening effect of partial local free entropy.
To this purpose, one can define an averaged loss over an $N$-dimensional vicinity in weight space --this imitating the effects of 
local entropy-- or to a lower-dimensional vicinity -- this instead imitating partial local entropy.
We focus on the latter case and define
\begin{equation}\label{simpler}
 \bar{\cal L}^{[\boldsymbol U]}(R, k, y; \boldsymbol W) \equiv 
 \frac{1}{y+1}\left[{\cal L}_{\text{c.e.}}(\boldsymbol  W)
 + \sum_{a=1}^y {\cal L}_{\text{c.e.}}(\boldsymbol  W+\Delta \boldsymbol W^a) K_{R,k}^{[\boldsymbol U]}(\Delta \boldsymbol W^a)\right]\ .
\end{equation}
Considering the $k\rightarrow \infty$ limit one obtains
\begin{equation}
 \bar{\cal L}^{[\boldsymbol U]}(R, y; \boldsymbol W) \equiv 
 \frac{1}{y+1}\left[{\cal L}_{\text{c.e.}}(\boldsymbol  W)
 + \sum_{a=1}^y {\cal L}_{\text{c.e.}}(\boldsymbol  W+\Delta^{[\boldsymbol U]} \boldsymbol W^a) \right]\ ,
\end{equation}
where $\Delta^{[\boldsymbol U]}$ means simply that the random vectors are sampled uniformly within the 
hyper-cube $H_{\boldsymbol W, R}^{[\boldsymbol U]}$ centered in $\boldsymbol W$ and extending 
along the direction indicated by the vector $\boldsymbol U$, its edges being $2R$ long.
In the limit of infinite samples, we have 
\begin{equation}
 \bar {\cal L}^{[\boldsymbol U]}(R; \boldsymbol W) \xrightarrow[y\rightarrow +\infty]{}\ \int_{H_{\boldsymbol W, R}^{[\boldsymbol U]}} 
 d^N W'\ {\cal L}_{\text{c.e.}}(\boldsymbol  W') \ ,
\end{equation}
and the loss reduces to a simple local average along a subset of directions in weight space.%
\footnote{
The loss function defined in \eqref{simpler} can be related to the \emph{robust ensemble} studied in \cite{Baldassi_2016},
which in turns is similar to the \emph{elastic averaging} proposed in \cite{zhang2014deep}.
}

\section{Experiments with fully-connected networks}
\label{exp_MF}

The focus of the first group of experiments is on layer-wise partial entropy regularizations for multi-layer, fully-connected neural networks trained on image classification tasks.
Namely, we considered partial local entropies where the subset of weights chosen for smoothening coincides with whole layers.
We consider the $10$-class classification tasks associated with MNIST \cite{726791} and 
Fashion-MNIST \cite{xiao2017/online} datasets, whose input images are $28$ pixels wide and $28$ pixels height.
We consider both 2-layer and 3-layer fully-connected neural networks with continuous 
weights%
\footnote{We performed the experiments with single floating point numerical precision.}
having a further $10$-neuron output layer. All layers except the last have $784=28^2$
neurons and are structurally identical, apart from their different depth within the network.
The following hyper-parameters have been kept fixed for all the experiments:
learning rate $\eta = 0.0001$, momentum $\mu=0.9$, mini-batch size $256$ and trained 
for $120$ epochs.

We considered two loss functions, a partial local exponential average loss (PLEA)
\begin{equation}\label{PLEA}
 {\cal L}_{\text{PLEA}}(\boldsymbol W) = -\log \left\{\frac{1}{1+y} \left[e^{-{\cal L}_{\text{c.e.}}(\boldsymbol W)} 
 + \sum_{a=1}^y  e^{-{\cal L}_{\text{c.e.}}(\boldsymbol W + \Delta \boldsymbol W^a)} \right] \right\}\ ,
\end{equation}
and a partial local average loss (PLA)
\begin{equation}\label{PLA}
 {\cal L}_{\text{PLA}}(\boldsymbol W) = \frac{1}{1+y} \left[{\cal L}_{\text{c.e.}}(\boldsymbol W) 
 + \sum_{a=1}^y  {\cal L}_{\text{c.e.}}(\boldsymbol W + \Delta \boldsymbol W^a) \right] \ ,
\end{equation}
where ${\cal L}_{\text{c.e.}}$ is the cross-entropy loss and $\Delta \boldsymbol W^a$ is a random 
vector sampled in a vicinity of $\boldsymbol W$.%
\footnote{The losses \eqref{PLEA} and \eqref{PLA} correspond to infinite sharpness limits, $k\rightarrow\infty$,
of \eqref{smu_U} and \eqref{simpler}, respectively. See Section \ref{PLE} for more details.}
Such a vicinity is a hyper-cube centered in $\boldsymbol W$ with edge $2R$ and extending only
along a subspace of the $N$-dimensional weight space.
Notice that in this way the regularizations of the cross-entropy ${\cal L}_{\text{c.e.}}$
given by \eqref{PLEA} and \eqref{PLA} enforce an anisotropic bias.

In the experiments reported below we consider only subspaces
associated to one or more layers at a time.%
\footnote{Throughout the present paper the weight space spanned by $\boldsymbol W$ is formed only by 
the synaptic coefficients connecting different layers, while it excludes biases. 
Despite these latter are present and trained over, we do not smooth over them.}
Apart from the entropic smoothening, we do not enforce any further regularization, 
in particular we do not use weight decay.

\newpage
\subsection{Results}

The experiments suggest two main conclusions:
\begin{itemize}
 \item In general, the entropic regularizations \eqref{PLEA} and \eqref{PLA}
 improve test accuracy. The effect increases rapidly with the size $R$ of the 
 smoothening region, up to a maximum size beyond which performance gets degraded.
 
 \item When implemented on suitable subsets of weights (\emph{e.g.} single layers), 
 the entropic regularizations outperform significantly their isotropic counterparts. 
 
\end{itemize}

The first point means that smoothening improves performance up to 
a point beyond which its averaging effect distorts the original loss landscape too heavily.
The second point means that the strong differences in the role played by the various weights 
affect the loss landscape and the effectiveness of regularization.
This implies that the shape of the wide flat minima encountered by SGD optimization is relevant, 
not only their extension.
Another generic conclusion suggested by the experiments is that the layer-wise entropic 
regularization is more effective when performed on deeper levels. This harmonizes with 
the intuitive idea that deeper weights are associated to more complex features,
which --in a reliable classification-- should be progressively more robust.

An important detail of the experimental setups is that all layers have the same number
of neurons, $784$. Thus, when comparing quantities associated to different layers, we are 
actually probing the mere effect of depth. A direct comparison between structurally 
different layers would instead be more difficult to interpret.

\subsection{2-layer fully-connected neural network on Fashion-MNIST}

We considered 2-layer, fully-connected neural networks adopting both
PLEA loss function \eqref{PLEA} and PLE loss function \eqref{PLA}.
The results obtained with the two loss functions are qualitatively analogous.

\begin{figure}
 \begin{center}
  \includegraphics[width=0.98\textwidth]{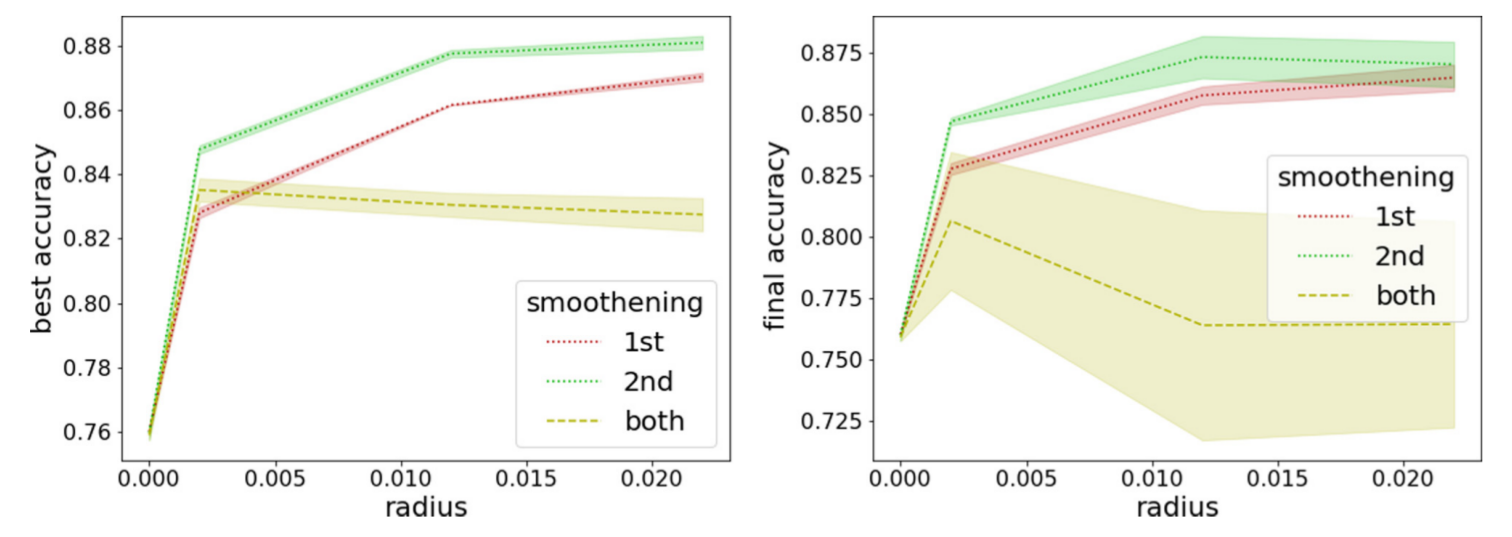}
 \end{center}
 \caption{Comparison of the best (left) and final (right) test accuracy reached by a 2-layer fully-connected neural network on 
 Fashion-MNIST. The lines correspond to three different PLA losses (see \eqref{PLA}) obtained 
 by smoothening the cross entropy respectively on the 1st, the 2nd or both layers.} 
 \label{acc_com}
\end{figure}

We measured the test accuracy reached by three versions of the same 2-layer network
as we moved the regularization radius $R$%
\footnote{\emph{i.e.} the parameter encoding the linear size of the smoothening region;
see Section \ref{PLE} for details.}, the three versions differ simply by the choice 
of the weight subspace considered for smoothening: 
either (i) the whole first layer; 
(ii) the whole second layer; 
(iii) both layers (isotropic choice).
The results are reported in Figure \ref{acc_com} and Figure \ref{acc_com_ENT} (left plot).
Regularization on the 2nd layer alone proved to be the best strategy for both 
choices of loss functions and in the entire range of $R$ probed by the experiments.
The isotropic regularization can outperform the regularization on the 1st layer 
alone, but only at very small values for $R$. In fact, the isotropic choice leads 
soon to degraded results as $R$ increases, while the single-layer regularizations 
continue to improve the test accuracy, showing a saturating behavior.

\subsection{3-layer fully-connected neural network on MNIST}

\begin{figure}
 \begin{center}
  \includegraphics[width=0.98\textwidth]{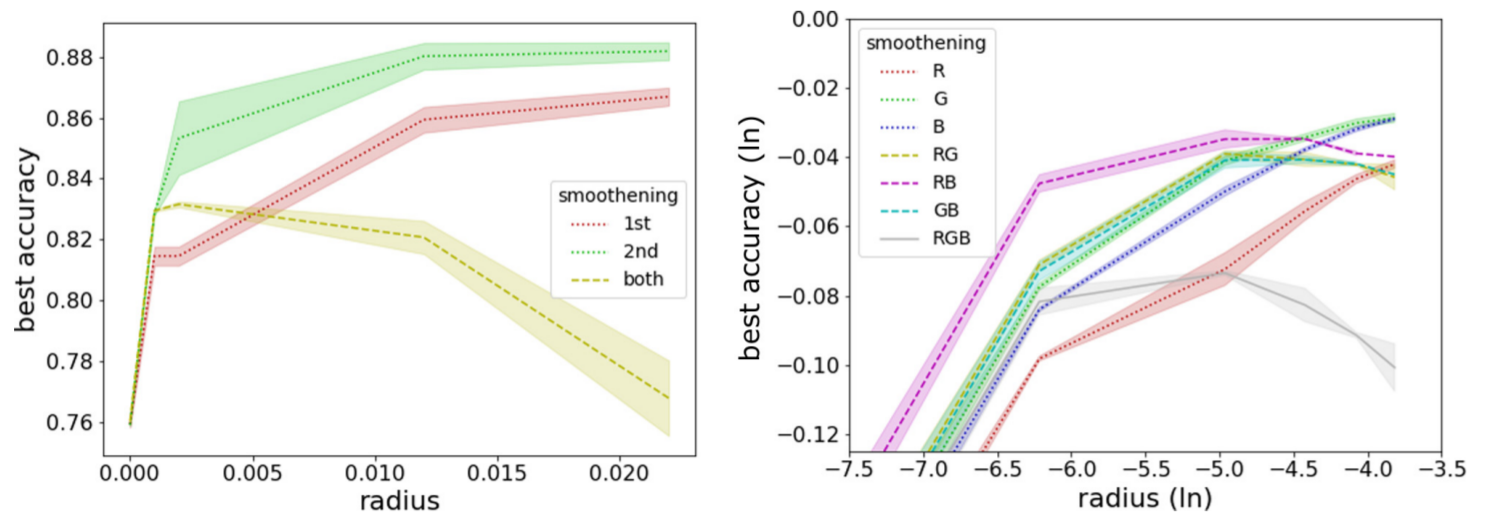}
 \end{center}
 \caption{Left plot: best test accuracy reached during training by a 2-layer fully-connected neural network over Fashion-MNIST.
 The three lines correspond to three different PLEA regularization schemes (see Eq.\eqref{PLE} where 
 smoothening is performed on the 1st layer alone, on the 2nd layer alone or on both layers, respectively.
 Right plot: best test accuracy reached by a 3-layer fully-connected neural network on MNIST. The lines represents different 
 PLA regularization schemes according to an RGB color nomenclature where Red corresponds to the 1st layer,
 Green to the 2nd and Blue to the 3rd.}
 \label{acc_com_ENT}
\end{figure}

The experiments on the 3-layer fully-connected neural networks confirm and extend the results obtained for its 2-layer counterpart.
They are depicted in \ref{acc_com_ENT} (right plot).
In particular, the isotropic choice proves to be the worst among all the possible choices of subsets%
\footnote{Recall that we consider only subsets of weights associated to one or more whole layers.}
as soon as the smoothening radius $R$ is sufficiently big. Moreover, there is an articulated interplay 
of regimes as $R$ varies: at the lowest values of $R$ the best choice consists in regularizing with respect 
to the 1st and 3rd layers jointly; at large values of $R$, regularizing with respect to the 2nd or 3rd layer 
alone proves to be the best choice. Also the performance hierarchy among the sub-optimal regularization schemes changes as $R$ 
moves showing a complicated structure.

\subsection{Finer sampling}
\label{ItS}

\begin{figure}
 \begin{center}
  \includegraphics[width=0.98\textwidth]{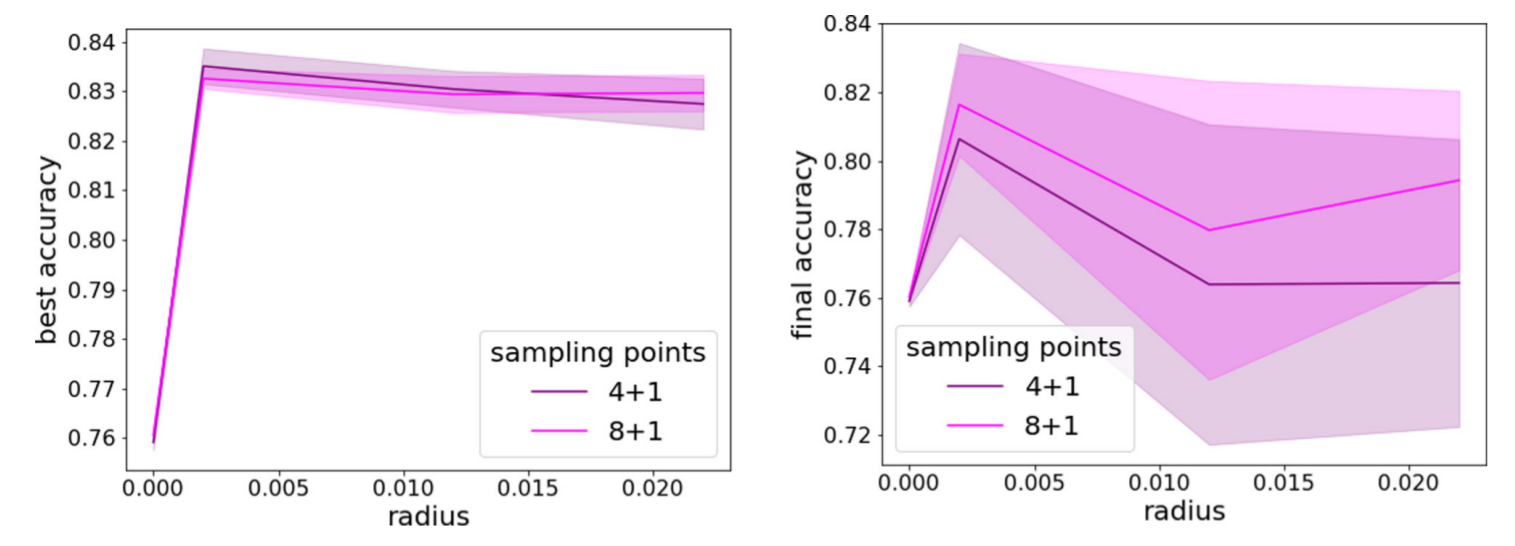}
 \end{center}
 \caption{Comparison of the test accuracy performance obtained with a bi-layer fully-connected neural network on Fashion-MNIST
 and trained with PLA loss (see Eq.\eqref{PLA}). The lighter line refer to finer sampling, $y=8$,
 while the darker line refers to $y=4$. There is no strong sensitivity to the sample size.}
 \label{48}
\end{figure}
In order to test whether the decrease in accuracy associated to regularizing on multiple layers is due to
insufficient sampling (\emph{i.e.} too small $y$, see \eqref{PLEA} and \eqref{PLA})
we repeated the experiments performed with the 2-layer fully-connected neural network on Fashion-MNIST with PLA loss doubling 
the number of sampling points $y$. The results obtained with $y=8$ are comparable to those obtained with 
$y=4$, see Figure \ref{48}; this hints to the fact that the sampling of the smoothening neighborhood 
can not explain the poor performance of multi-layer regularization.

\section{Experiments with convolutional neural networks}

We extend the analysis described in Section \ref{exp_MF} in two directions: i) we consider more complicated image classification tasks;
ii) we consider deeper networks with convolutional architectures. The convolutional neural networks that we adopt present 5 
convolutional layers followed by 3 fully-connected layers, the detail of the architectures are given in \eqref{conv_cifar} and \eqref{conv_stl}.
All the convolutional kernels are $3\times 3$.

\subsection{Results}

The main conclusions which emerged from the study of partial entropic regularizations applied to convolutional neural networks are the following:
\begin{enumerate}
 \item Partial entropic regularizations involving the convolutional layers lead, in general, to worse classification accuracy with respect to the non-regularized case;
 \item When applied to the fully-connected head of convolutional networks, partial entropic regularizations can improve the classification accuracy 
 in a similar way as observed on multi-layered perceptrons, described in Section \ref{exp_MF}. Especially when adopting an \emph{early stopping}
 strategy interrupting training before its full stabilization.
\end{enumerate}

Some further comments are in order. Convolutional layers implement a structured bias
encoding some degree of \emph{locality} and \emph{translational invariance}. 
Thus the convolutional structure, if compared to fully-connected layers, is highly specialized.
Entropic regularizations can in general be thought of as corresponding to the integration over some injected artificial noise,
as such, one expects them to weakens, if not even to spoil, any specific bias previously encoded in the neural architecture.
Such comment holds both for the partial entropic regularization studied here, as well as for other forms 
of noisy regularizations, like Dropout. This latter, too, has been observed to hamper the performance 
of convolutional networks \cite{devries2017improved}. Conversely, fully-connected layers have no specific structure
and the average over additional noise can lead to better performance in general, also when applied to 
the fully-connected head in a convolutional network.

In the experiments detailed below, we consider fully-connected heads formed by three layers.
The deepest layer outputs 10 channel, as required by the 10-class classification tasks considered
and we do not regularize it. The other two fully-connected layers are instead equal in shape among themselves. 
As already argued in Section \ref{exp_MF} for the multi-layer perceptrons, the structural equality allows for a direct comparison 
between the two layers.

\subsection{CIFAR10}

For the classification task corresponding to the CIFAR10 dataset we considered the following convolutional architecture:
\begin{equation}\label{conv_cifar}
 \begin{array}{c|c|c}
  \text{layer} & \text{in channels} & \text{out channels}\\
  \hline
  \text{Conv} & 3 & 64\\
  \text{Conv} & 64 & 64\\
  \hline
  \text{MaxPool} &&\\
  \hline
  \text{Conv} & 64 & 128\\
  \text{Conv} & 128 & 128\\
  \text{Conv} & 128 & 128\\
  \hline
  \text{MaxPool} &&\\
  \hline
  \text{Fully} & 128\cdot 4\cdot 4 & 128\cdot 4\cdot 4\\
  \text{Fully} & 128\cdot 4\cdot 4 & 128\cdot 4\cdot 4\\
  \text{Fully} & 128\cdot 4\cdot 4 & 10\\
  \hline
 \end{array}
\end{equation}
{We have trained it for 360 epochs with a constant learning rate $\eta=10^{-4}$, a mini-batch size of 256 images and momentum $\mu=0.9$
without Nesterov acceleration. The training dataset has been augmented/regulated by means of random transformations on the images,
specifically we have considered rescaled random crops ranging from $60\%$ to $100\%$ of the image area and with a height/width 
ratio form $\frac{3}{4}$ to $\frac{4}{3}$. Neither weight decay, nor dropout layers have been used.%
\footnote{We compare the partial entropic regularizations against weight-decay regularizations in Subsubsection \ref{wd}.}
Actually, the only regularization 
for the stochastic gradient descent has been provided by the partial local average, encoded in \eqref{PLA}, with 4 additional 
sample points drawn from a uniform distribution in hypercube ball of side equal to $2R$, with $R = 0.01$. The initialization followed the so-called 
Kaiming procedure described in \cite{he2015delving}.

The results are depicted in Figure \ref{CIGAR10}. We considered three cases: no regularization or PLA regularization applied to
either the first or second layer in the fully-connected head (see \eqref{conv_cifar}). The PLA modification of the loss function
yields better performance, both in-sample and out-of-sample, especially if combined with an early stopping strategy
which interrupts the training before its eventual stabilization. The PLA procedure implies collecting multiple samples of the loss function 
in the vicinity of the current weight configuration of the network (we took 4 points in a hypercubic vicinity plus the center); the gradient is 
accumulated but eventually rescaled in such a way that the multiple sampling does not affect the training by means of a simple amplification 
of the learning rate.
\begin{figure}[!h]
 \begin{center}
  \includegraphics[width=0.49\textwidth]{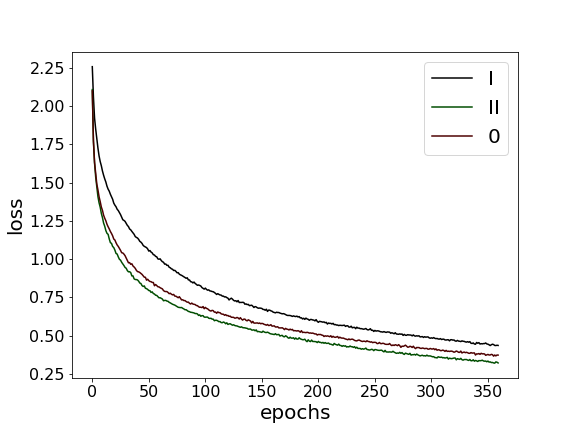}
  \includegraphics[width=0.49\textwidth]{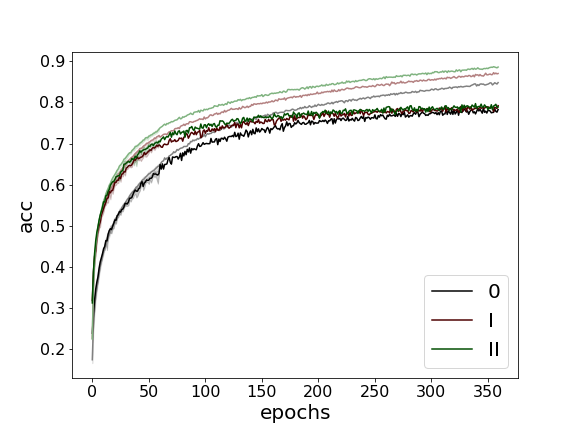}
 \end{center}
 \caption{Training loss (left) and test accuracy (right) of the convolutional network \eqref{conv_cifar} trained to solve the 
 CIFAR10 image classification task. The black lines refer to the case where no entropic regularization was considered. The 
 red and green lines refers to PLA (see \eqref{PLA}) loss applied to the first (I) or second (II) fully-connected layer, respectively,
 as specified in the legend. The paler lines in the right plot depict the in-sample accuracy.}
 \label{CIGAR10}
\end{figure}

\subsubsection{Comparison against standard weight-decay regularization}
\label{wd}

In order to better assess the effects of partial entropic regularization,
we considered comparing them with those produced by a standard regularization method,
namely, weight decay \cite{10.5555/2986916.2987033}. Specifically, we considered 
three levels of weight decay rate, 0.01, 0.001 and 0.0001. As shown in Figure \ref{wd_fig},
weight decay regularization proved to be of essentially no utility in the present experiments.
On the contrary, partial entropic regularization proved to improve the performance,
more significantly in the early phase of the training, only slightly in later stages.
These experiments do not pretend to support a generic claim, however they show explicitly 
that partial entropic regularization can be preferable with respect to weight regularization.
\begin{figure}[h!]
 \begin{center}
  \includegraphics[width=0.49\textwidth]{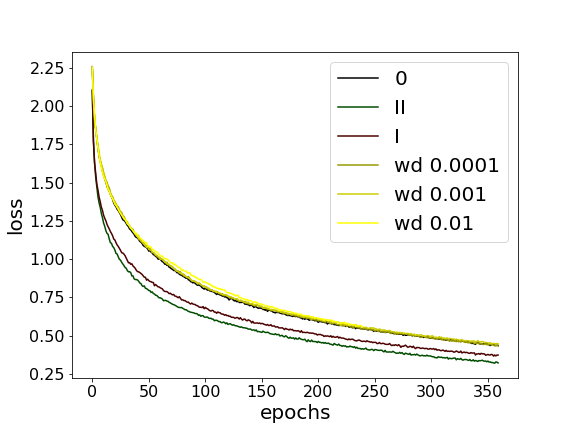}
  \includegraphics[width=0.49\textwidth]{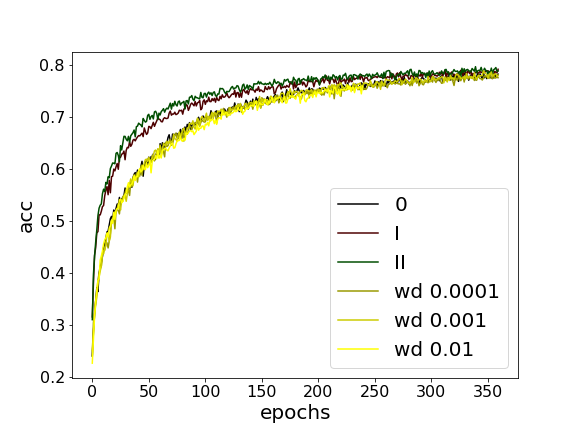}
  \includegraphics[width=0.49\textwidth]{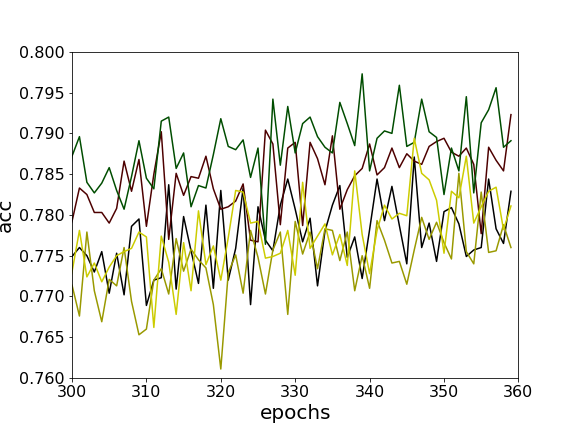}
 \end{center}
 \caption{The paler (yellow) lines represent training instances which adopted increasing levels of weight decay regularization,
 0.01, 0.001 and 0.0001, respectively from darker to lighter. They all overlap with the unregularized case (label 0), meaning that the effects 
 of weight regularization are irrelevant. 
 The plot on the top left depicts the in-sample loss during training on CIFAR10; the lowest line corresponds to II and the second from below corresponds to I.
 The plot on the top right shows the out-of-sample accuracy during training on CIFAR10; the uppermost line corresponds to II while the second from top corresponds to I. 
 The bottom plot is a zoom of the top right figure highlighting the late portion of the training.}
 \label{wd_fig}
\end{figure}

\subsection{STL10}

STL10 is a 10-class classification dataset of $96\times 96$ color images acquired from ImageNet.
STL10 was designed for partially unsupervised learning \cite{inproceedingsa}, in fact it contains only 500 labeled images for supervised training.
Although these hardly suffice to train a machine in a fully supervised setup, we use them to simply show the positive effects 
that partial entropy regularizations induce on the early phase of the training, without requiring an overall satisfactory performance.%
\footnote{STL10 has already been used in the literature for supervised learning, see for example \cite{weiler2019general}.}

We adopt the following convolutional architecture
\begin{equation}\label{conv_stl}
 \begin{array}{c|c|c}
  \text{layer} & \text{in channels} & \text{out channels}\\
  \hline
  \text{Conv} & 3 & 8\\
  \text{Conv} & 8 & 8\\
  \hline
  \text{MaxPool} &&\\
  \hline
  \text{Conv} & 8 & 16\\
  \text{Conv} & 16 & 16\\
  \text{Conv} & 16 & 16\\
  \hline
  \text{MaxPool} &&\\
  \hline
  \text{Fully} & 16\cdot 20\cdot 20 & 16\cdot 20\cdot 20\\
  \text{Fully} & 16\cdot 20\cdot 20 & 16\cdot 20\cdot 20\\
  \text{Fully} & 16\cdot 20\cdot 20 & 10\\
  \hline
 \end{array}
\end{equation}
which is analogous to \eqref{conv_cifar}, but has lighter layers.
We trained it for 960 epochs with a constant learning rate $\eta=10^{-5}$, momentum $\mu=0.9$ without Nesterov acceleration and a mini-batch size of $64$ images.
In order to mitigate the issue presented by the smallness of the training set, we have applied heavy augmentation and regularization to the training images.
Specifically, we considered random crops whose size ranges from $8\%$ to the full image, and whose aspect ratio ranges from $\frac{3}{4}$ to $\frac{4}{3}$;
we considered random horizontal flips, random reduction to gray-scale (with a probability $p=0.1$), color jitter (brightness, contrast, saturation 
and hue all set to $0.5$) and random rotation whose maximal rotation angle is $\pm\pi$ radiants.%
\footnote{To implement such transformations we relied on the \emph{transforms} library in PyTorch.}
\begin{figure}[h!]
 \begin{center}
  \includegraphics[width=0.49\textwidth]{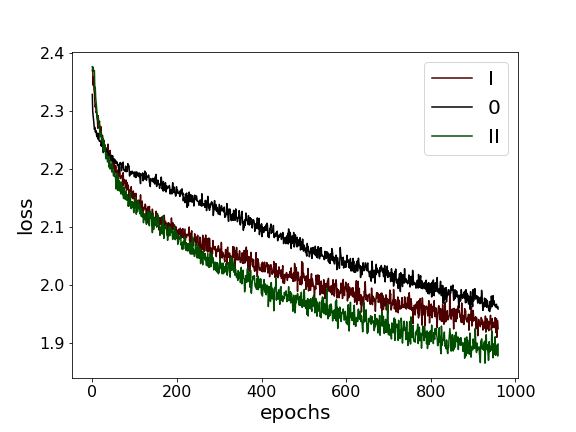}
  \includegraphics[width=0.49\textwidth]{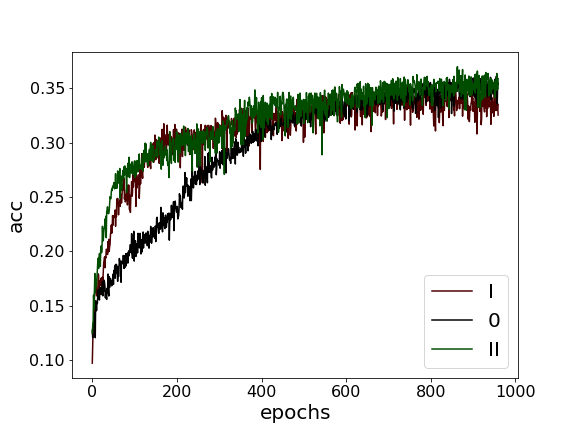}
  \includegraphics[width=0.49\textwidth]{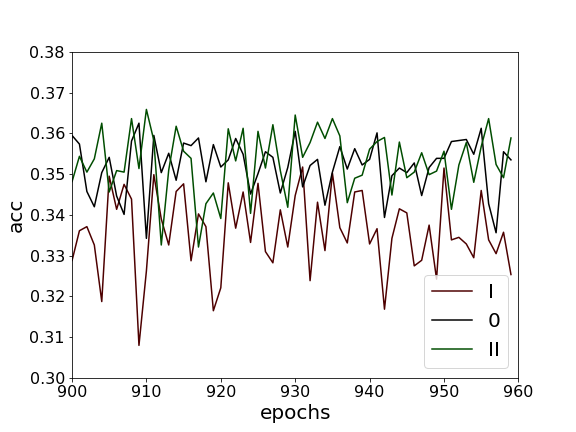}
 \end{center}
 \caption{Training loss (left) and test accuracy (right) of a deep convolutional network \eqref{conv_stl}
 trained on the STL10 image dataset in a fully supervised scheme for 960 epochs. The black lines correspond to zero entropic regularization,
 while the lines I and II correspond to partial entropic regularization applied only to the 1st and 2nd fully-connected layers, respectively.
 The green line (II) correspond to lowest loss and highest accuracy.
 The bottom plot is a zoom over the last part of the right plot above, depicting the accuracy levels reached at the end of the training.}
 \label{STL10}
\end{figure}

We monitored the training and report the evolution of the in-sample loss and the test accuracy in Figure \ref{STL10}.
The partial entropic regularization, applied to one layer at a time, improves the training and validation performances,
but only if accompanied with an early stopping strategy.
The experiments of Figure \ref{STL10} refer to a PLA loss \eqref{PLA} where the side of the sampled hypercube is $2R$ with $R=0.01$. 
The network was initialized according to the Kaiming method \cite{he2015delving}, no weight decay was considered.

\section{Discussion}

A local smoothening of the loss function can improve the chase for wide flat minima \cite{Baldassi2015SubdominantDC,2016arXiv161101838C},
which is already a strength of the standard stochastic gradient descent algorithm \cite{Xie2020ADT}.%
\footnote{The generic relevance of \emph{wide flat minima} is still debated in the literature \cite{liao2020generalization,Poggio30039},
especially in relation to scale covariance and normalization in weight space for networks adopting ReLU activations.}
We elaborate and refine the smoothening techniques based on local entropy to the purpose of leveraging the anisotropic nature of deep weight spaces.
Concretely, we propose to restrict local entropic losses to suitable sub-spaces of weights, thus defining \emph{partial local entropies}.
This allows us to explore, address and exploit the intrinsic anisotropic nature of deep weight spaces.
In fact, we show that a partial entropic regularization can implement useful biases
on the shape of the minima encountered by SGD optimization.

We have mainly explored the layer-wise implementations of partial local entropies; although there is room 
for finer analyses resolving smaller sub-spaces, the layer-wise approach is both natural (\emph{i.e.}
well-adapted to the architecture of deep networks) and informative.

In the present paper we have applied partial entropic regularizations to some
fully-connected and convolutional neural networks employed for image classification tasks, 
they can however be employed for the optimization of wider classes
of learning machines, \emph{e.g.} auto-encoders \cite{negri2019natural}. In particular, the specific layer-wise 
entropic regularizations proposed in the present study apply in any context involving a layered neural network.
The partial entropic regularizations have been proved to be potentially useful in all the considered experiments.
However, their positive effects in progressively more demanding tasks seem to be restricted to an early stopping protocol.
The adoption of a partial entropic loss led to a more aggressive optimization in all the performed experiments.

\subsection{Direct analysis in the language of statistical physics}

The study of local entropic regularizations is a very active research front in machine learning, 
especially in connection to statistical physics 
\cite{zhang2014deep,Baldassi2015SubdominantDC,chaudhari2016entropysgd,Baldassi_2016,Baldassi2016LocalEA,Baldassi_2019,negri2019natural,baldassi2019clustering,pittorino2020entropic}. 
Wide flat minima have been described as a structural 
characteristic of deep networks and their correlation with good generalization performance has been 
claimed in \cite{Baldassi2015SubdominantDC,Baldassi_2019}. In some simple setups, it is even possible to estimate analytically 
the hyper-volume of the clusters of configurations giving rise to the relevant minima \cite{PhysRevLett.65.2312,Baldassi_2019}.
The theoretical framework on which the calculations are based has been developed for the study of 
disordered systems in condensed matter, mainly spin glasses (see \cite{mezard1987spin} and references therein), 
it is called \emph{replica approach}. Within this approach, different regimes are described by different ansatzes
and can be separated by clustering transitions \cite{Castellani_2005}.%
\footnote{An analogous transition in $K$-SAT problems has been studied in \cite{Mezard812,Krzakaa10318}.}

A simple version of the replica approach \cite{mezard2009information} can rely on two (crude) assumptions:
(i) averaging over (typically Gaussian) input; (ii) considering tree-like architectures. 
The former essentially washes out completely the information about the dataset.
This is not always undesirable, in fact it allows for the characterization of structural 
properties of the machines that hold true \emph{per se} independently of the dataset. It however 
constitutes a limitation whenever the actual information provided by the input is important. 
As a future prospect, it would be interesting to study how a direct and explicit account
of correlations in the input data could improve the theoretical understanding of the partial entropic regularizations,
especially regarding their effects on the inference quality.%
\footnote{The study performed in \cite{Kabashima_2008} is relevant to this purpose, 
nevertheless it would require a generalization beyond single-layer networks.}
Considering a tree-like architecture is very helpful to simplify the computations, in fact 
avoiding loops in the network often opens the possibility of exact computations by, for instance, 
belief propagation algorithms \cite{mezard2009information,Baldassi_2019}. Nevertheless,
adopting a tree-like network as a proxy for a fully-connected one can be too crude a simplification
which is expected to deviate more significantly as the depth of the system is increased.

In order to explain the experiments described in Section \ref{exp_MF}, it would be desirable to have a direct 
control on the shape of the relevant clusters of weight configurations reached upon SGD training, or at least an 
estimation thereof. This could be seen as a refinement of the estimation of the clusters size \cite{PhysRevLett.65.2312,Baldassi_2019},
as such it is likely to be a very demanding endeavor up to the point that it becomes natural to ask 
whether some simpler --though possibly rougher-- approach is viable. 
To this purpose, it is interesting to investigate mean-field inference methods \cite{Gabri2020}.

\section{Acknowledgements}

I would like to thank Riccardo Argurio, Diogo Buarque Franzosi, Stefano Goria,
Javier M\'as, Andrea Mezzalira, Giorgio Musso, Alfonso Ramallo, Hern\'an Serrano and Maurice Weiler
for interesting discussions.

\bibliographystyle{plain}
\bibliography{PLE}

\begin{thebibliography}{10}

\bibitem{Baldassi_2016}
Carlo Baldassi, Christian Borgs, Jennifer~T. Chayes, Alessandro Ingrosso, Carlo
  Lucibello, Luca Saglietti, and Riccardo Zecchina.
\newblock Unreasonable effectiveness of learning neural networks: From
  accessible states and robust ensembles to basic algorithmic schemes.
\newblock {\em Proceedings of the National Academy of Sciences},
  113(48):E7655–E7662, Nov 2016.

\bibitem{Baldassi2015SubdominantDC}
Carlo Baldassi, Alessandro Ingrosso, Carlo Lucibello, Luca Saglietti, and
  Riccardo Zecchina.
\newblock Subdominant dense clusters allow for simple learning and high
  computational performance in neural networks with discrete synapses.
\newblock {\em Physical review letters}, 115 12:128101, 2015.

\bibitem{Baldassi2016LocalEA}
Carlo Baldassi, Alessandro Ingrosso, Carlo Lucibello, Luca Saglietti, and
  Riccardo Zecchina.
\newblock Local entropy as a measure for sampling solutions in constraint
  satisfaction problems.
\newblock {\em Journal of Statistical Mechanics: Theory and Experiment},
  2016:023301, 2016.

\bibitem{Baldassi_2019}
Carlo Baldassi, Enrico~M. Malatesta, and Riccardo Zecchina.
\newblock Properties of the geometry of solutions and capacity of multilayer
  neural networks with rectified linear unit activations.
\newblock {\em Physical Review Letters}, 123(17), Oct 2019.

\bibitem{baldassi2019clustering}
Carlo Baldassi, Riccardo~Della Vecchia, Carlo Lucibello, and Riccardo Zecchina.
\newblock Clustering of solutions in the symmetric binary perceptron, 2019.

\bibitem{PhysRevLett.65.2312}
E.~Barkai, D.~Hansel, and I.~Kanter.
\newblock Statistical mechanics of a multilayered neural network.
\newblock {\em Phys. Rev. Lett.}, 65:2312--2315, Oct 1990.

\bibitem{Castellani_2005}
Tommaso Castellani and Andrea Cavagna.
\newblock Spin-glass theory for pedestrians.
\newblock {\em Journal of Statistical Mechanics: Theory and Experiment},
  2005(05):P05012, May 2005.

\bibitem{2016arXiv161101838C}
Pratik {Chaudhari}, Anna {Choromanska}, Stefano {Soatto}, Yann {LeCun}, Carlo
  {Baldassi}, Christian {Borgs}, Jennifer {Chayes}, Levent {Sagun}, and
  Riccardo {Zecchina}.
\newblock {Entropy-SGD: Biasing Gradient Descent Into Wide Valleys}.
\newblock {\em arXiv e-prints}, page arXiv:1611.01838, November 2016.

\bibitem{chaudhari2016entropysgd}
Pratik Chaudhari, Anna Choromanska, Stefano Soatto, Yann LeCun, Carlo Baldassi,
  Christian Borgs, Jennifer Chayes, Levent Sagun, and Riccardo Zecchina.
\newblock Entropy-sgd: Biasing gradient descent into wide valleys, 2016.

\bibitem{DBLP:journals/corr/abs-1710-11029}
Pratik Chaudhari and Stefano Soatto.
\newblock Stochastic gradient descent performs variational inference, converges
  to limit cycles for deep networks.
\newblock {\em CoRR}, abs/1710.11029, 2017.

\bibitem{inproceedingsa}
Adam Coates, Honglak Lee, and Andrew Ng.
\newblock An analysis of single-layer networks in unsupervised feature
  learning.
\newblock pages 1--9, 01 2011.

\bibitem{DASILVA200416}
P.C. [da Silva], L.R. [da Silva], E.K. Lenzi, R.S. Mendes, and L.C. Malacarne.
\newblock Anomalous diffusion and anisotropic nonlinear fokker–planck
  equation.
\newblock {\em Physica A: Statistical Mechanics and its Applications},
  342(1):16 -- 21, 2004.
\newblock Proceedings of the VIII Latin American Workshop on Nonlinear
  Phenomena.

\bibitem{devries2017improved}
Terrance DeVries and Graham~W. Taylor.
\newblock Improved regularization of convolutional neural networks with cutout,
  2017.

\bibitem{Wei}
Weinan E.
\newblock A proposal on machine learning via dynamical systems.
\newblock {\em Communications in Mathematics and Statistics}, 5:1--11, 02 2017.

\bibitem{Gabri2020}
Marylou Gabrié.
\newblock Mean-field inference methods for neural networks.
\newblock {\em Journal of Physics A: Mathematical and Theoretical},
  53(22):223002, May 2020.

\bibitem{goldfeld2019estimating}
Ziv Goldfeld, Ewout van~den Berg, Kristjan Greenewald, Igor Melnyk, Nam Nguyen,
  Brian Kingsbury, and Yury Polyanskiy.
\newblock Estimating information flow in deep neural networks, 2019.

\bibitem{he2015delving}
Kaiming He, Xiangyu Zhang, Shaoqing Ren, and Jian Sun.
\newblock Delving deep into rectifiers: Surpassing human-level performance on
  imagenet classification, 2015.

\bibitem{Kabashima_2008}
Y~Kabashima.
\newblock Inference from correlated patterns: a unified theory for perceptron
  learning and linear vector channels.
\newblock {\em Journal of Physics: Conference Series}, 95:012001, Jan 2008.

\bibitem{10.5555/2986916.2987033}
Anders Krogh and John~A. Hertz.
\newblock A simple weight decay can improve generalization.
\newblock In {\em Proceedings of the 4th International Conference on Neural
  Information Processing Systems}, NIPS'91, page 950–957, San Francisco, CA,
  USA, 1991. Morgan Kaufmann Publishers Inc.

\bibitem{Krzakaa10318}
Florent Krzaka{\l}a, Andrea Montanari, Federico Ricci-Tersenghi, Guilhem
  Semerjian, and Lenka Zdeborov{\'a}.
\newblock Gibbs states and the set of solutions of random constraint
  satisfaction problems.
\newblock {\em Proceedings of the National Academy of Sciences},
  104(25):10318--10323, 2007.

\bibitem{726791}
Y.~{Lecun}, L.~{Bottou}, Y.~{Bengio}, and P.~{Haffner}.
\newblock Gradient-based learning applied to document recognition.
\newblock {\em Proceedings of the IEEE}, 86(11):2278--2324, Nov 1998.

\bibitem{liao2020generalization}
Qianli Liao, Brando Miranda, Lorenzo Rosasco, Andrzej Banburski, Robert Liang,
  Jack Hidary, and Tomaso Poggio.
\newblock Generalization puzzles in deep networks, 2020.

\bibitem{mezard2009information}
M.~M{\'e}zard and A.~Montanari.
\newblock {\em Information, Physics, and Computation}.
\newblock Oxford Graduate Texts. OUP Oxford, 2009.

\bibitem{mezard1987spin}
M.~Mezard, G.~Parisi, and M.A. Virasoro.
\newblock {\em Spin Glass Theory And Beyond: An Introduction To The Replica
  Method And Its Applications}.
\newblock World Scientific Lecture Notes In Physics. World Scientific
  Publishing Company, 1987.

\bibitem{Mezard812}
M.~M{\'e}zard, G.~Parisi, and R.~Zecchina.
\newblock Analytic and algorithmic solution of random satisfiability problems.
\newblock {\em Science}, 297(5582):812--815, 2002.

\bibitem{Musso:2020itr}
Daniele Musso.
\newblock {Stochastic gradient descent with random learning rate}.
\newblock 3 2020.

\bibitem{negri2019natural}
Matteo Negri, Davide Bergamini, Carlo Baldassi, Riccardo Zecchina, and
  Christoph Feinauer.
\newblock Natural representation of composite data with replicated
  autoencoders, 2019.

\bibitem{pittorino2020entropic}
Fabrizio Pittorino, Carlo Lucibello, Christoph Feinauer, Enrico~M. Malatesta,
  Gabriele Perugini, Carlo Baldassi, Matteo Negri, Elizaveta Demyanenko, and
  Riccardo Zecchina.
\newblock Entropic gradient descent algorithms and wide flat minima, 2020.

\bibitem{Poggio30039}
Tomaso Poggio, Andrzej Banburski, and Qianli Liao.
\newblock Theoretical issues in deep networks.
\newblock {\em Proceedings of the National Academy of Sciences},
  117(48):30039--30045, 2020.

\bibitem{Poggio}
Maximilian Riesenhuber and Tomaso Poggio.
\newblock Riesenhuber, m. and poggio, t. hierarchical models of object
  recognition in cortex. nat. neurosci. 2, 1019-1025.
\newblock {\em Nature neuroscience}, 2:1019--25, 12 1999.

\bibitem{michael2018on}
Andrew~Michael Saxe, Yamini Bansal, Joel Dapello, Madhu Advani, Artemy
  Kolchinsky, Brendan~Daniel Tracey, and David~Daniel Cox.
\newblock On the information bottleneck theory of deep learning.
\newblock In {\em International Conference on Learning Representations}, 2018.

\bibitem{sharma2020neural}
Utkarsh Sharma and Jared Kaplan.
\newblock A neural scaling law from the dimension of the data manifold, 2020.

\bibitem{DBLP:journals/corr/Shwartz-ZivT17}
Ravid Shwartz{-}Ziv and Naftali Tishby.
\newblock Opening the black box of deep neural networks via information.
\newblock {\em CoRR}, abs/1703.00810, 2017.

\bibitem{2017arXiv171006451S}
Samuel~L. {Smith} and Quoc~V. {Le}.
\newblock {A Bayesian Perspective on Generalization and Stochastic Gradient
  Descent}.
\newblock {\em arXiv e-prints}, page arXiv:1710.06451, October 2017.

\bibitem{weiler2019general}
Maurice Weiler and Gabriele Cesa.
\newblock General $e(2)$-equivariant steerable cnns, 2019.

\bibitem{xiao2017/online}
Han Xiao, Kashif Rasul, and Roland Vollgraf.
\newblock Fashion-mnist: a novel image dataset for benchmarking machine
  learning algorithms, 2017.

\bibitem{Xie2020ADT}
Zeke Xie, Issei Sato, and Masashi Sugiyama.
\newblock A diffusion theory for deep learning dynamics: Stochastic gradient
  descent escapes from sharp minima exponentially fast.
\newblock {\em ArXiv}, abs/2002.03495, 2020.

\bibitem{zhang2014deep}
Sixin Zhang, Anna Choromanska, and Yann LeCun.
\newblock Deep learning with elastic averaging sgd, 2014.

\bibitem{2018arXiv180300195Z}
Zhanxing {Zhu}, Jingfeng {Wu}, Bing {Yu}, Lei {Wu}, and Jinwen {Ma}.
\newblock {The Anisotropic Noise in Stochastic Gradient Descent: Its Behavior
  of Escaping from Sharp Minima and Regularization Effects}.
\newblock {\em arXiv e-prints}, page arXiv:1803.00195, February 2018.

\end{thebibliography}

\end{document}